\documentclass{article}

\usepackage{arxiv}

\usepackage[utf8]{inputenc} 
\usepackage[T1]{fontenc}    
\usepackage{hyperref}       
\usepackage{url}            
\usepackage{booktabs}       
\usepackage{amsfonts}       
\usepackage{nicefrac}       
\usepackage{microtype}      
\usepackage{lipsum}		
\usepackage{graphicx}
\usepackage{amssymb}
\usepackage{amsmath}  
\usepackage{bm}
\usepackage{textcomp}

\title{Rethinking Fully Convolutional Networks for the Analysis of Photoluminescence Wafer Images}

\author{{\hspace{1mm}Maike Lorena Stern}\thanks{corresponding author} \\
	Osram Opto Semiconductors\\
	Leibnizstraße 4\\
	93055 Regensburg, Germany \\
	\\
	Department of Computer Science \\
	Friedrich Alexander University Erlangen-Nuremberg\\
	91058 Erlangen, Germany \\
	\texttt{maike.stern@osram-os.com} \\
	\And
	{\hspace{1mm}Hans Lindberg} \\
	Osram Opto Semiconductors\\
	Leibnizstraße 4\\
	93055 Regensburg, Germany \\
	\And
	{\hspace{1mm}Klaus Meyer-Wegener} \\
	Department of Computer Science \\
	Friedrich Alexander University Erlangen-Nuremberg\\
	Martensstraße 3\\
	91058 Erlangen, Germany 
}

\renewcommand{\headeright}{Preprint}
\renewcommand{\undertitle}{Preprint}

\begin{document}
\maketitle

\begin{abstract}
	The manufacturing of light-emitting diodes is a complex semiconductor-manufacturing process, interspersed with different measurements. Among the employed measurements, photoluminescence imaging has several advantages, namely being a non-destructive, fast and thus cost-effective measurement. On a photoluminescence measurement image of an LED wafer, every pixel corresponds to an LED chip's brightness after photo-excitation, revealing chip performance information. However, generating a chip-fine defect map of the LED wafer, based on photoluminescence images, proves challenging for multiple reasons: on the one hand, the measured brightness values vary from image to image, in addition to local spots of differing brightness. On the other hand, certain defect structures may assume multiple shapes, sizes and brightness gradients, where salient brightness values may correspond to defective LED chips, measurement artefacts or non-defective structures. In this work, we revisit the creation of chip-fine defect maps using fully convolutional networks and show that the problem of segmenting objects at multiple scales can be improved by the incorporation of densely connected convolutional blocks and atrous spatial pyramid pooling modules. We also share implementation details and our experiences with training networks with small datasets of measurement images. The proposed architecture significantly improves the segmentation accuracy of highly variable defect structures over our previous version. 
\end{abstract}

%
%


\section{Introduction}
\label{Intro}

\begin{figure*}
\centering
\includegraphics[scale=.4]{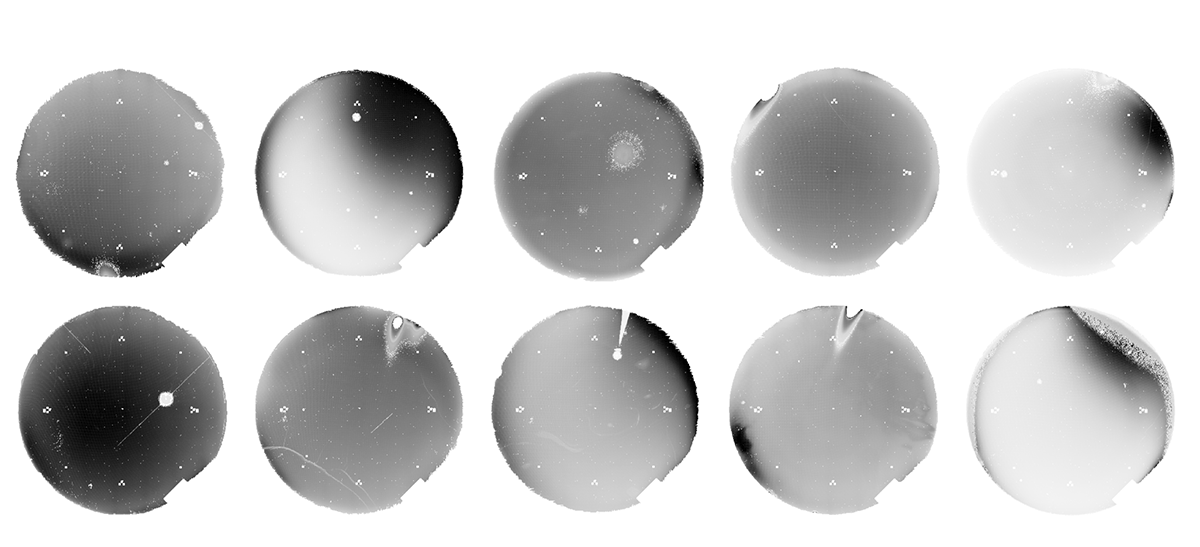}
\caption{Overview over the variety of photoluminescence measurement results with different brightness values as well as defect cluster shapes and sizes. Due to the measurement setup brightness values can differ significantly within a wafer as well as from wafer to wafer. Additionally, defect clusters assume different sizes and shapes along with an interchange of sharp and smooth brightness gradients.}
\label{fig:multiplewafers}
\end{figure*}

In the manufacturing of light-emitting diodes (LEDs), measurements constitute an inevitable but simultaneously unwanted process step, given that they add no value to the product but rather enable the monitoring of product and process. Because the manufacturing of LEDs is a complex semiconductor-manufacturing process, it includes a variety of different measurements, employed for process monitoring, the determination of LED-chip properties and the detection of conspicuous or defective LED chips. Based on these measurements, defective LED chips can be rejected directly instead of being processed further. Among the available measurement methods photoluminescence imaging has several advantages, namely being a non-destructive, fast and thus cost-effective measurement. Because it is also a non-contact measurement, photoluminescence measurements can be performed early in the manufacturing process, before electrical contact pads are added to the chip surface, and can also be applied to advanced chip designs without a contact pad. By irradiating the surface of an LED wafer photoluminescence is induced, ultimately causing the emission of photons. Comparing the recorded optical intensity image with chip-fine electrical and optical measurements reveals that almost all defective LED chips manifest as salient brightness values. However, not all salient brightness values are defective LED chips but may also correspond to functional or non-defect structures, such as film tears that arise from an earlier process step. Additionally, photoluminescence measurements generate images with varying distributions of brightness values from wafer to wafer as well as in local areas, as shown in figure \ref{fig:multiplewafers}. Moreover, possible defect structures cover multiple scales, from single defective chips to elongated cracks, voids and defect clusters.  

Here, defect clusters appear the fewest of all defect structures and assume various shapes, sizes and brightness gradients, which exacerbates an accurate segmentation. As a result, a pattern recognition algorithm that maps a photoluminescence image to a defect map must be able to reliably distinguish between salient brightness values corresponding to defective LED chips, measurement artefacts and good chips, respectively, and in addition accurately segment multiple scaled objects.

In this work, we revisit the creation of defect maps from photoluminescence images, using fully convolutional networks \cite{Long}. In a previous study  \cite{Stern}, a specifically designed encoder-decoder fully-convolutional-network architecture was introduced that enabled a chip-fine output resolution, correctly recognised non-defect structures and measurement artefacts and accurately segmented common defect structures. Not only appear these defect structures frequently in the dataset, their appearance remains relatively uniform, as well. Defect clusters, however, appear comparatively rarely and in variable shapes, sizes and with different brightness gradients, as shown in figure \ref{fig:multiplewafers}. Because the procurement of a dataset from a running production is limited by the manufacturing process, the selection of photoluminescence images with defect clusters is not representative of the true distribution. As a result, defect clusters with a very uncommon shape or size were often segmented inaccurately.

Therefore, we study the incorporation of additional architectural concepts to the network architecture, namely densely connected convolutional blocks, abbreviated as dense blocks \cite{Huang} as well as atrous spatial pyramid pooling (ASPP) modules \cite{Chen}. Here, dense blocks input all preceding, concatenated feature maps, which improves the information flow through the network and results in implicit deep supervision \cite{Lee}. Moreover, as noted by \cite{Huang} and observed in our experiments, dense network designs have a regularising effect that diminishes overfitting, which is beneficial if only small datasets are available for training. ASPP modules, on the other hand, address the challenge of multiple scaled objects by probing incoming feature maps with different field-of-views at once and additionally incorporate image-level context via global average pooling \cite{Liu}. Our experiments show that for the special composition of photoluminescence images the combination of two ASPP modules yields the highest network performance, with one module in the encoder (downsampling) and one module in the decoder (upsampling) path. Altogether, the dense ASPP$^2$ version of our model increases test defect-class accuracy from 83.5\,\% to 91.5\,\% and test mean pixel accuracy from 94.1\,\% to 96.7\,\%, trained on a dataset of 111 photoluminescence wafer images and 25 validation images. Additionally, the segmentation accuracy of uniquely shaped defect clusters improves empirically, revealing that the careful design of fully convolutional networks can compensate for comparably small datasets.

\section{Related Work}

Fully convolutional networks for semantic segmentation \cite{ZeilerDec,Long,Lin,Zhao,Chen,Zhou} have become powerful tools in a data scientist's toolbox, given that they enable object localisation on a pixel-level without the need of handcrafted image-processing pipelines. Their application covers a range of tasks, especially with respect to everyday-scene images \cite{Segnet,Liu,Pohlen,Tiramisu,Chen2, SABOKROU201888} and medical images \cite{Ronneberger,Tai,Tran,Sharma,Guan}. Note that to the best of our knowledge, no other working groups have reported on fully-convolutional-network architectures for the segmentation of photoluminescence wafer images. 

\textbf{Dense Blocks:} For the segmentation of everyday-scene images, it has been shown that the incorporation of dense blocks enables condensed, parameter-efficient network architectures, which diminish overfitting and strengthen feature propagation. The concept of dense blocks has first been introduced by \cite{Huang} for convolutional neural networks  \cite{MNIST,Simonyan,Szegedy}. Hereby, the idea of residual networks \cite{ResNets} is advanced by connecting each network layer to every other layer and thereby increasing the number of direct connections from $ L $ to $ \frac{L(L+1)}{2} $ for a network with $ L $ layers. Since their introduction, the concept of dense blocks has also been adapted to fully-convolutional-network architectures \cite{Tiramisu, Zhu2, Arsalan, Khened}. 

\textbf{Atrous Spatial Pyramid Pooling:} ASPP modules \cite{Lazebnik,Grauman,Zhao,Chen}, on the other hand, address the challenge arising from differently scaled objects, such as single defective LED chips and defect clusters. For this purpose, ASPP modules arrange several layers in parallel, which probe the incoming feature maps at multiple rates, using atrous (also called dilated) convolutions \cite{Holschneider, Sermanet, Yu, Chen, Hamaguchi}. Moreover, an additional layer captures global context on an image-level so as to maintain consistency in the labelling of distant pixels \cite{Liu, Zhao}. By evaluating these diverse feature information simultaneously, ASPP modules refine the segmentation accuracy of multi-scaled objects.

\section{Data}
\begin{figure}[h]
\centering
\includegraphics[width=.5\linewidth]{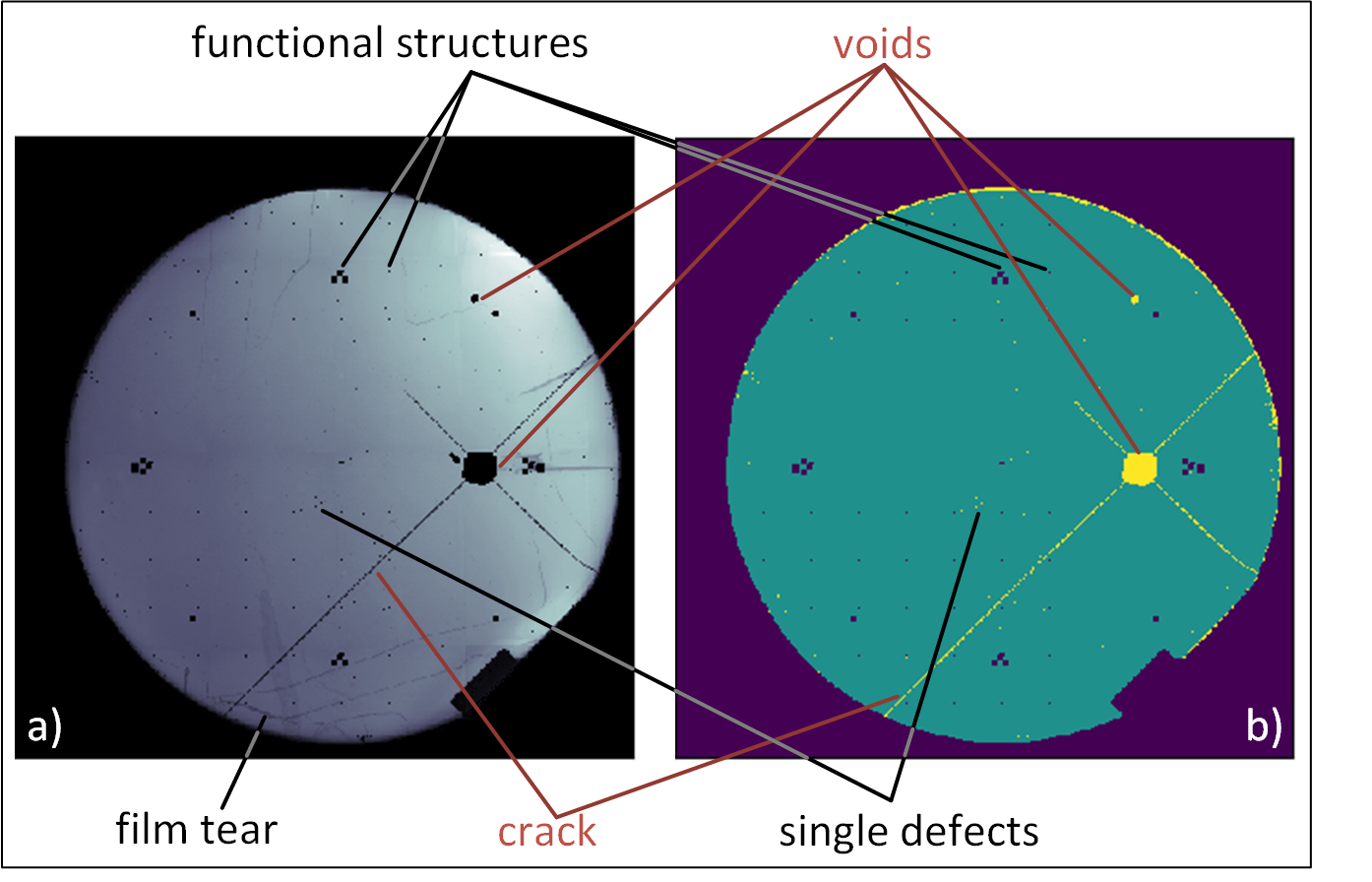}
\caption{Photoluminescence input image of a wafer with 133,717 LED chips~(a) and the corresponding wafer probing-based label image (b), with yellow = defective chips, turquoise = in-spec chips and blue = functional structures/background. We classify defect structures by their appearance, namely single defects, cracks, voids and defect clusters. Note that film tears are visible on the photoluminescence image but not the label image, because they do not correlate to defective LED chips. The photoluminescence image has been embedded with functional structures and additional evaluation results, in order to align input and label images.}
\label{fig:pl-probing}
\end{figure}

Datasets for semantic segmentation depend on the possibility to obtain pixel-wise labels. In our case, manual labelling is impossible because the depicted wafers consist of 133,717~LED chips each, which must be classified chip-wise. Therefore, we use the results of a comprehensive electrical and optical measurement, named wafer probing, which takes place late in the manufacturing process when electrical contact pads are already added to the chip. However, using measurement results as both, input and label images, may result in input-label mismatches, caused by different measurement techniques. On the one hand, the brightness determined by photoluminescence measurement differs from the brightness determined by wafer probing, where electronic excitation is induced rather than photo-excitation. And on the other hand, wafer probing results cover a compilation of several electrical and optical tests, including evaluations that take place before wafer probing, such as ultrasonic measurements. To align input and label images, we proceeded as follows:

\textbf{Input Images:} Photoluminescence-measurement results are not saved as an image but as a list in a text file, where a brightness value is reported for each LED chip over an 8\,bit greyscale. In order to create a photoluminescence image, a zero matrix of size $ 442 \times 440 $ is filled with chip values, where applicable. To reach accordance with the wafer prober-based label images, the aforementioned ultrasonic-measurement results are embedded into the photoluminescence images, in addition to functional structures, which are applied only after the measurement (see figure \ref{fig:pl-probing}). 

\textbf{Label Images:} As mentioned, wafer probing covers a sequence of tests and because the first failure that occurs during the probing sequence is assigned as defect cause, defect classes do not necessarily correspond with the core reason for failure. Additionally, defect causes might have been assigned based on evaluations that took place before wafer probing. Therefore, we subsumed all defect causes into one class (figure~\ref{fig:pl-probing}, yellow) and created two additional classes, one representing in-spec chips (turquoise) and the other one representing background as well as functional structures (dark blue), and yield $ 442 \times 440 \times 3 $ label images. Note that not all defects determined by wafer probing manifest in salient brightness values and thus a small number of input-label mismatches remains.  

To train the network, we split our dataset of 136 images into 111 training and 25 validation images. Even though it is generally agreed upon that the success of deep-learning methods can be attributed to large-scale labelled datasets \cite{Sun}, in our case the number of training images is limited by the rare occurrence of wafers with salient and unique defect structures in the manufacturing process. We therefore employ data augmentation by rotating the images 45°, 90° and 135°, respectively, and hereby quadruple the number of training samples. To test the developed network architecture, we employ a test dataset of 366~photoluminescence images taken from the running production, where only samples with distinctive differences between input image and label image as well as fractured wafers were discarded.

\section{Network Architecture}

\begin{figure} [htb!]
\centering
\includegraphics[width=1.\linewidth]{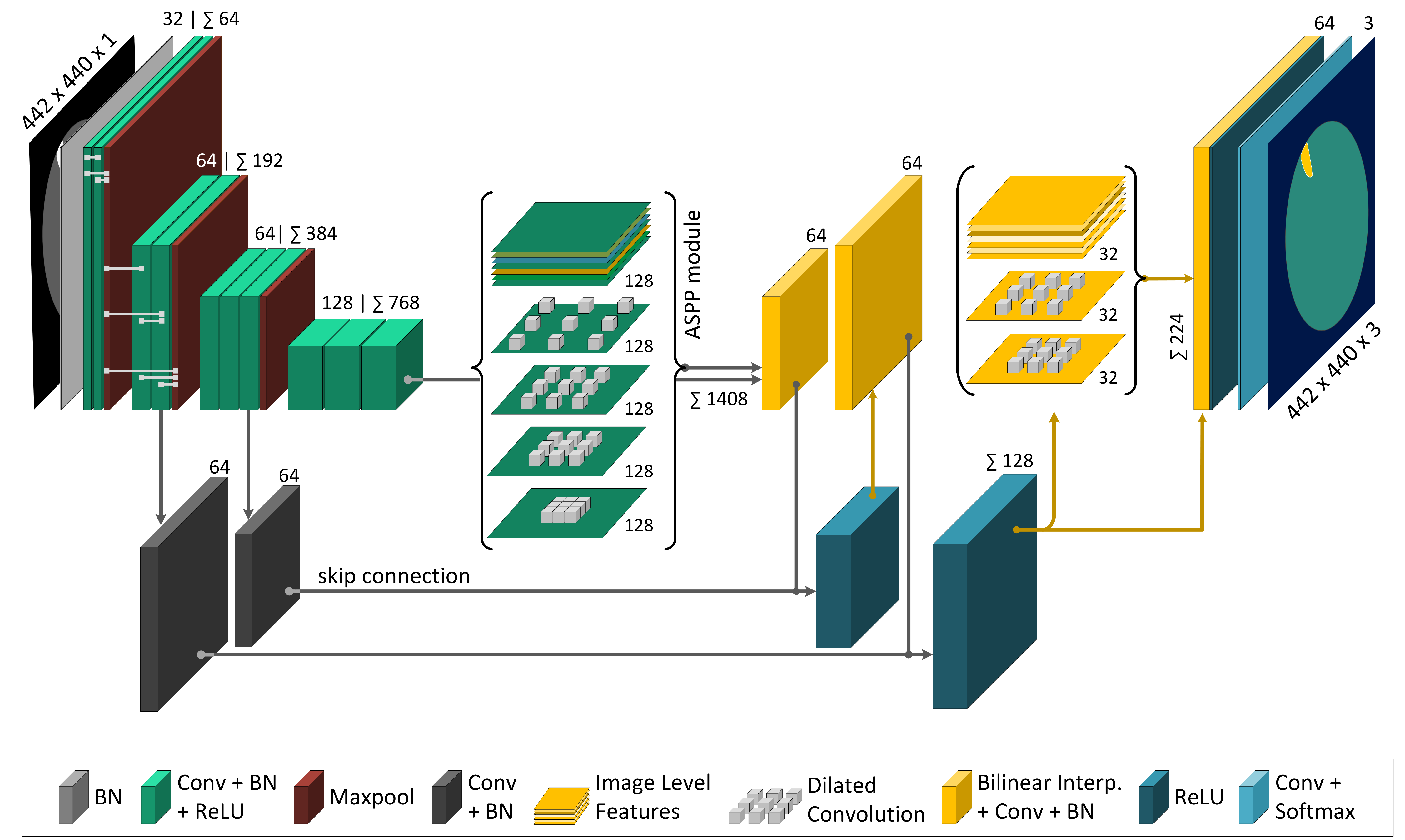}
\caption{Visualisation of the developed dense architecture with two ASPP modules, where the legend states the layer operations and BN equals batch normalisation. The numbers on top of each dense block and layer, respectively, denote the number of kernels per layer and the number of concatenated output feature maps per block (indicated by a sum sign), if applicable. The horizontal grey lines in the first two blocks illustrate the dense connections, exemplarily for all downsampling blocks. ASPP modules are indicated by brackets, where each module layer processes the same input simultaneously. Layers below the main branch indicate skip connections, which concatenate feature maps of shallow and deep layers.}
\label{fig:neueArch}
\end{figure}

Network architectures for industrial tasks are designed to be deployed in the manufacturing process, making as few as possible pre-processing steps beneficial. The developed network architecture, depicted in figure~\ref{fig:neueArch}, therefore inputs unnormalised photoluminescence images and the first network layer applies a learned normalisation to the input image \cite{Ioffe}. Following batch normalisation, the now normalised image is forwarded to the first dense convolutional pooling block, which covers two convolutional layers with 32 kernels each (figure~\ref{fig:neueArch}, green blocks) and a subsequent maxpooling operation (red block). All convolutional layers in the downsampling path apply a composite function $ H_l $ to the incoming feature maps, consisting of the three consecutive operations $ 3 \times 3 $~convolution, batch normalisation and ReLU activation function, where the input feature maps are zero-padded so as to keep their resolution stable. To implement dense connections, the maxpooling operation inputs the concatenated feature maps of both preceding layers and reduces the incoming tensor dimensions to $ 221 \times 220 \times 64 $ with a $ 2 \times 2 $ kernel and a stride of 2, before forwarding them to the subsequent block. 

The horizontal grey lines added to the first two blocks in figure~\ref{fig:neueArch} visualise the dense connections, where the number of kernels per convolutional layer as well as the number of output feature maps of each block is written on top of the corresponding block (marked with a sum sign). As an instance, the second block employs two convolutional layers with 64~kernels each and the first layer inputs the concatenated, maxpooled 64 feature maps of the previous block, with $ x_l = H_l([x_{0}, ..., x_{l-1}]) $, where $ x_l $ are the feature maps of the current layer $ l $, $ H_l $ is the aforementioned composite function and $ [x_{0}, ..., x_{l-1}] $ denotes the concatenation of the previous feature maps. Subsequently, a maxpooling layer downsamples and forwards all 192~feature maps to the third block. Because dense connections encourage the reuse of feature maps, the number of kernels in each downsampling layer was reduced, as shown in table \ref{tab:dense}. This is especially beneficial for very large input images (which occur frequently in manufacturing) because it allows shallow networks without loss of accuracy rather than processing a cropped and stitched image. 

\begin{table}[h]
\centering
\small
\caption{Comparison of a basic network version with a typical number of kernels \cite{Stern} and a densely connected version with a reduced number of kernels in each layer. Note that the number of encoder feature maps increases in the dense model because the bypassed and upsampled feature maps are concatenated rather than added.}
\label{tab:dense}       
\begin{tabular}{lll}
	\hline\noalign{\smallskip}
	& basic model  & dense model \\
	\noalign{\smallskip}\hline\noalign{\smallskip}
	conv1\_x & 64, 64        & 32, 32 \\
	conv2\_x & 128, 128      & 64, 64 \\
	conv3\_x & 256, 256, 256 & 64, 64, 64 \\
	conv4\_x & 512, 512, 512 & 128, 128, 128 \\
	conv5\_x & 512, 512, 64  & 128, 128, 128 \\
	up1 & 64 & 128 \\
	up2 & 64 & 128\\
	up3 & 64 & 128 \\
	up4 & 3  & 3  \\
	\noalign{\smallskip}\hline
	\noalign{\smallskip}
\end{tabular}
\end{table}

Altogether, the network's encoder (or downsampling path) consists of four dense blocks, which repeatedly decrease feature-map resolution while increasing the number of kernels. Even though the network's receptive field is widened by this typical bi-pyramid design, all image objects are analysed with the same receptive field, despite their multiple sizes and shapes. Therefore, an ASPP module completes the downsampling path, which inputs the concatenated, un-pooled 768 feature maps of the fourth block and processes them in parallel. The module itself consists of four dilated-convolution layers and one global-average-pooling layer, which extracts image-level features. Here, each of the dilated-convolution layers samples the incoming feature maps with another dilation rate~$ r $, with $ z_{i',j',f'} =  \sum_{i=1}^{H} \sum_{j=1}^{W} x_{i'+ir-1,j'+jr-1,f} w_{ijff'}$, where $ z_{i',j',f'} $ is a pixel in the output feature map $ \bm{f}' $ and $ \bm{w} $ is a $ H \times W $ sized filter that is slid across the sparsely sampled input feature map~$ \bm{x}$. As shown in the following section, the best results have been achieved with $ r = $ 1, 2, 6, and 12, where a dilation rate of 2 employed with a $ 3 \times 3 $ kernel increases the convolved area to $ 5 \times 5 $ pixels, where only every other pixel is used. Incidentally, all dilated-convolution layers apply the consecutive operations dilated convolution, batch normalisation and ReLU activation function. The fifth module layer prevents fragmented segmentation results by incorporating image-level features. For this purpose, global average pooling is applied, where the incoming $ 56 \times 55 \times 768 $ feature maps are downsampled to $ 1 \times 1 \times 768 $, that is one feature per feature map is extracted. To retrieve the previous dimensions and decrease the number of feature maps to 128, the consecutive operations $ 1 \times 1 $~convolution, batch normalisation, bilinear upsampling and ReLU activation function are applied. Finally, the module output and input are concatenated, resulting in 1408 feature maps altogether, and forwarded to the first upsampling layer. 

As shown in figure \ref{fig:neueArch}, the decoder (or upsampling path) uses skip connections to refine the network's output by combining the upsampled coarse semantic information with fine-grain local information from shallow layers. For this purpose, the first upsampling layer inputs the aforementioned 1408 feature maps and outputs 64 upsampled feature maps. Here, all upsampling layers apply the consecutive operations bilinear interpolation, $ 3 \times 3 $~convolution and batch normalisation. After upsampling, the resulting feature maps are concatenated with the bypassed feature maps of the skip connection, whose number of feature maps has first been reduced to 64 using $ 3 \times 3 $ convolution and batch normalisation. Finally, a ReLU activation function is applied to the concatenated 128 feature maps, which are then forwarded to the second upsampling layer. Note that the upsampling layers are not actually densely connected but rather input dense blocks via the skip connections, because the consistently increasing feature-map dimensions would otherwise require unreasonable computational resources \cite{Tiramisu, Zhu2}. 

The second upsampling layer consists of two parts, the aforementioned upsampling procedure as well as a second ASPP module. Altogether, the module consists of three parallel layers, which all output 32 feature maps, namely two dilated-convolution layers with $ r = 2 $ and 4 and a global-average-pooling layer. Then, the concatenated module input and output is forwarded to the final network block, which calculates the three output maps in two steps: first, the incoming feature maps are upsampled as before. Hereby, the network pre-processes the feature maps from the preceding layer and restores the original dimensions of the input image. Then, the number of feature maps is reduced to three and the normalised probability values of each output pixel are determined, using $ 1 \times 1 $ convolution followed by a softmax activation function. If the network output is used for classification rather than network training, a subsequent \textit{argmax} function may be employed to determine the most probable class category for each image pixel. To train the network for 80 epochs, we used a decaying learning rate of $ 5 \cdot 10^{-4} $ as well as an L2-regularisation strength of $ 5 \cdot 10^{-4} $ in combination with an RMSprop optimiser \cite{Hinton}. Furthermore, to equalise the unbalanced class categories, we added a loss weight of 2,000 to defect-class losses and a weight of 100 on the remaining two classes. Additionally, we initialised the first four layers with pre-trained VGG\,16 weights \cite{Simonyan} via transfer learning \cite{Torrey}, where we had to adjust the number of kernels of the transferred parameters first. 

\section{Experimental Evaluation}

\begin{figure}[htb!]
\centering
\includegraphics[width=.7\linewidth]{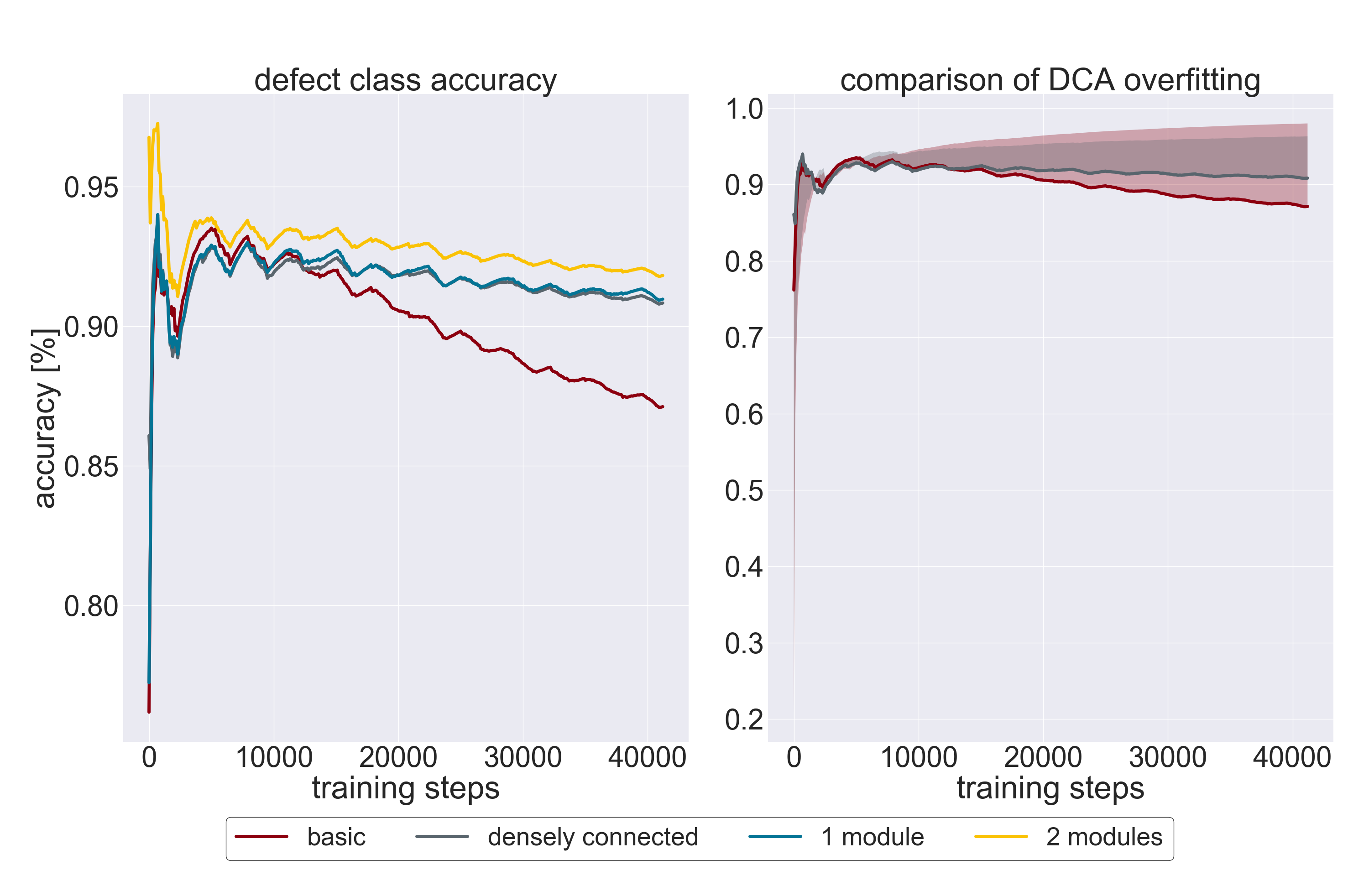}
\caption{Left: Defect class accuracy of a basic encoder-decoder architecture (red) compared to the a dense version (grey) as well as a dense model with one (blue) and two atrous-spatial-pyramid-pooling modules (\textit{dense ASPP²}, yellow). The right panel visualises the varying divergence of validation (solid line) and training defect-class accuracy (area over the line) of the basic and the dense models.}
\label{fig:advancedarchitectures}
\end{figure}

To evaluate the effect of the introduced architectural concepts---dense blocks and ASPP modules---five models are compared: the basic model introduced in \cite{Stern}, a dense version of this model as well as three dense versions with additional ASPP modules, namely one module in the downsampling path, one module in the upsampling path and one model with both modules (denoted as \textit{dense ASPP²}). Figure \ref{fig:advancedarchitectures} (left) depicts the validation defect-class accuracy of four architectures, where "1~module" refers to the version with an ASPP module in the downsampling path. 

Analysing the performance of the basic version (red line) reveals substantial overfitting, which is also verified by figure \ref{fig:advancedarchitectures}, right plot, where the difference between training and validation defect-class accuracy is visualised as area over the line for the basic version (red) and the dense version (grey). The comparison shows that introducing dense connections while reducing the number of kernels per layer in the downsampling path (see table \ref{tab:dense}) diminishes overfitting and thus increases defect-class accuracy as well as mean pixel accuracy, as listed in table \ref{tab:testArch}. 

\begin{table}[h]
\centering
\small
\caption{Validation and test mean pixel accuracy (MPA) and defect-class accuracy (DCA) of five different network models, where the \textit{dense ASPP²} model yields the highest overall performance. The test results reveal a drop in test defect-class accuracy for all five models, where the \textit{dense ASPP²} version shows the slightest difference in test and validation performance.}
\label{tab:testArch}       
\begin{tabular}{l|ll|ll}
	\hline\noalign{\smallskip}
	& MPA & test & DCA  & test \\
	\noalign{\smallskip}\hline\noalign{\smallskip}
	basic model             & 94.3 & 94.1 & 86.6 & 83.5 \\
	dense model             & 95.6 & 95.7 & 90.6 & 88.7 \\
	1 module (encoder)      & 95.5 & 96.1 & 91.0 & 90.4 \\
	dense ASPP²             & \textbf{96.2} & \textbf{96.7} & \textbf{92.0} & \textbf{91.5} \\
	1 module (decoder)      & 95.4 & 95.8 & 91.2 & 88.5 \\
	\noalign{\smallskip}\hline
\end{tabular}
\end{table}

After implementing dense connections to the network architecture, we experimented with ASPP modules and noticed that introducing an ASPP module into the network's encoder increases defect-class accuracy only slightly, while mean pixel accuracy remains roughly the same, as listed in table \ref{tab:testArch}. However, adding an additional ASPP module to the network's decoder further increased defect-class accuracy to 92.0\,\% as well as mean pixel accuracy to 96.2\,\%. Incidentally, adding a single ASPP module to the decoder yields comparable results as one module in the network's encoder, which indicates that for the given task probing incoming feature maps of a different size with multiple rates at once extracts complementing information. This assumption is supported by the models' test performances: while test mean pixel accuracy remains relatively stable, test defect-class accuracy drops for all five models, but shows the slightest difference for the \textit{dense ASPP²} model. Thus, adding two ASPP modules to the network architecture optimises the network's ability to generalise to previously unseen defect structures and increases segmentation accuracy.

\begin{figure}[htb!]
\centering
\includegraphics[width=.8\linewidth]{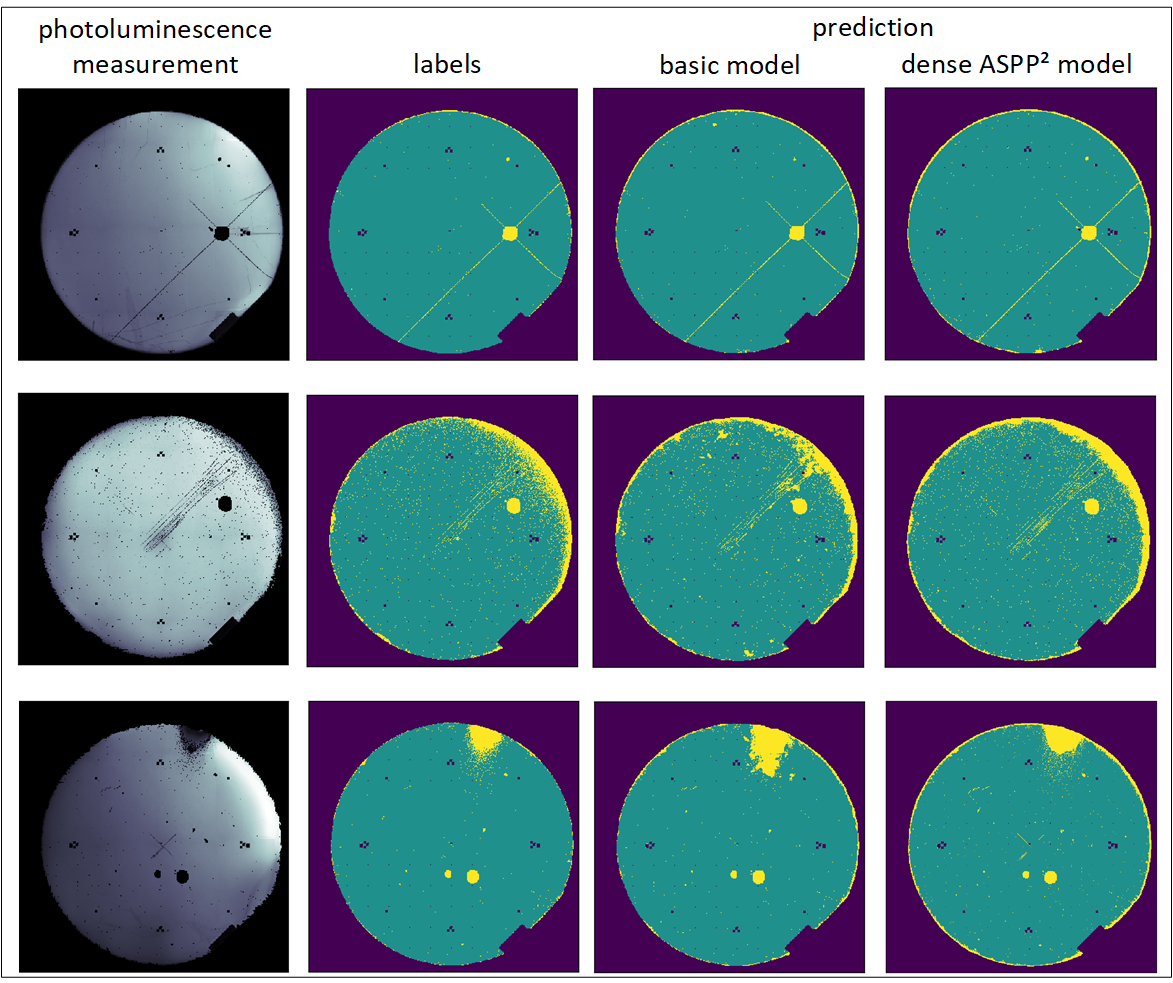}
\caption{Comparison of prediction images of a basic network architecture and the \textit{dense ASPP²} model, which exhibits the improved segmentation accuracy of the latter network design.}
\label{fig:vergleichArch}
\end{figure}

We can verify the \textit{dense ASPP²} network's increased segmentation accuracy by analysing prediction images of both networks, the basic model and the \textit{dense ASPP²} model, as shown in figure \ref{fig:vergleichArch}. Comparing the prediction images in the first row reveals that both architectures achieve comparable results for common defect structures, such as single defective chips, cracks and voids and have also learned to distinguish salient brightness values correlating to defective chips from those correlating to measurement artefacts, film tears and functional structures. However, the second and third row visualise the improved segmentation accuracy of the \textit{dense ASPP²} model, where unique structures, namely areas with dense single defective chips as well as large defect clusters, are segmented more precisely.

Finally, we examined the employed dilation rate in both single-module versions, the one in the encoder and the one in the decoder, as listed in table \ref{tab:testR}. It becomes apparent that for the given task a moderate increase in the dilation rate yields the highest performance, in contrast to \cite{Chen}, where a module with a large increase in the dilation rate, with $ r $ = 1, 6, 12, 18, achieved the best results. This result may be attributed to the different image objects of the used datasets, given that the latter architecture was developed for everyday-scene images. Here, image objects range from a few pixels to objects that cover almost the whole picture \cite{COCO}, whereas even large defect structures occupy only fractures of a wafer and thus comparably small image areas. As a result, the differences in object size vary less distinctively on photoluminescence images, which reflects in the optimal dilation rates of the ASPP modules. 

\begin{table}[h]
\centering
\small
\caption{Validation defect-class accuracy of different dilation rates in the ASPP modules, incorporated in the encoder and decoder structure, respectively. Note that for the second module a dilation rate of $ r = 1 $ is covered by the corresponding upsampling layer.}
\label{tab:testR}       
\begin{tabular}{ll|ll}
	\hline\noalign{\smallskip}
	\multicolumn{2}{c}{encoder} & \multicolumn{2}{c}{decoder}\\
	\hline\noalign{\smallskip}
	dilation rate &  DCA & dilation rate &  DCA  \\
	\noalign{\smallskip}\hline\noalign{\smallskip}
	1, 2, 4, 8   & 90.1  & 2, 1  & 89.5 \\
	1, 2, 6, 12  & \textbf{91.0}  & 2, 4  & \textbf{91.2} \\
	1, 4, 8, 16  & 90.1  & 4, 8  & 88.4 \\
	1, 6, 12, 18 & 89.0  & 6, 12 & 89.4 \\		 
	\noalign{\smallskip}\hline
\end{tabular}
\end{table}

\section{Conclusion}

In this paper, we have extended our previous network architecture for the pixel-fine analysis of photoluminescence wafer images with densely connected convolutional layers and two atrous-spatial-pyramid-pooling modules, one in the encoder and one in the decoder structure. Hereby, challenges arising from very small datasets could be addressed: dense blocks diminish overfitting and thus yield a higher defect-class accuracy, while ASPP modules refine segmentation accuracy by probing incoming feature maps at different scales at once. The resulting \textit{dense ASPP²} model increases test mean pixel accuracy from 94.1\,\% to 96.7\,\% and test defect-class accuracy from 83.5\,\% to 91.5\,\% and empirically improved the segmentation of unique defect structures, compared to our previous version.

\section*{Acknowledgments}
We would like to acknowledge support from the German Federal Ministry of Education and Research (BMBF), as part of the joint project \textit{InteGreat}. Moreover, we would like to thank those who share their knowledge generously and in particular the authors of NumPy groupies.

\bibliographystyle{unsrt}
\bibliography{references}  

\end{document}